\documentclass{article}
\usepackage[final,dblblindworkshop]{neurips_2026}
\workshoptitle{CODEC-FM: Collaborative, Open, and DECentralized training of Foundation Models}
\usepackage[utf8]{inputenc}
\usepackage[T1]{fontenc}

\usepackage{hyperref}
\usepackage{url}
\usepackage{booktabs}
\usepackage{amsfonts}
\usepackage{amsmath}
\usepackage{graphicx}
\usepackage{microtype}
\usepackage{multicol}
\usepackage{multirow}
\usepackage{xcolor}
\usepackage{placeins}
\newcommand{\mlpaux}{MLP-aux}
\newcommand{\tfaux}{Transformer-aux}
\newcommand{\tfmtp}{Transformer-aux\,+\,MTP}

\title{Greedy Local Learning for Language Model Pretraining:
Gaps and Objective Design}

\author{%
  Jihwan Moon\thanks{Work done while at KC ML2.}\\
  INFOCZ Inc.\\
  Seoul National University\\
  \texttt{jihwan.moon@vision.snu.ac.kr}\\
  \And
  Sheir A.~Zaheer\\
  KC ML2\\
  \texttt{sheir@kc-ml2.com}\\
  \And
  Jinmyoung Lee\\
  KC ML2\\
  \texttt{mys1458@kc-ml2.com}\\
  \AND
  Gunhee Kim\\
  Seoul National University\\
  \texttt{gunhee@snu.ac.kr}\\
  \And
  Chan Y.~Park\\
  KC ML2\\
  \texttt{chan.y.park@kc-ml2.com}\\
}

\begin{document}

\maketitle

\begin{abstract}
Greedy block-wise local learning splits a network into gradient-isolated blocks, each trained by its own auxiliary loss, so no backward pass crosses a block boundary. Inter-stage communication then becomes forward-only and every block can step its optimizer independently, which makes the approach directly relevant to decentralized model-parallel training. 
Local learning is competitive with end-to-end backpropagation on image classification, and on small Transformers it is known to trade a worse best loss for parallel speedup. 
How this loss gap behaves in autoregressive language model (LM) pretraining at larger scale, and which auxiliary designs reduce it, has to our knowledge not been measured. 
We present a token-budget-matched empirical study at 125M and 400M parameters with $K\in\{1,2,4\}$ blocks at Chinchilla-optimal budgets, factorizing the auxiliary design into network architecture and training objective. 
We observe that 
(i) the gap to end-to-end training more than doubles from $K{=}2$ to $K{=}4$, but at $K{=}4$ shrinks from 125M to 400M; 
(ii) replacing an MLP auxiliary with a Transformer-based one, a strong network-side intervention, recovers 22--41\% of the gap;
(iii) a multi-token-prediction (MTP) auxiliary objective helps at the first block boundary but hurts when added at deeper ones, and restricting it to the first block yields the best $K{=}4$ configuration ($+0.062$ vs.\ $+0.075$ nats at 400M); 
and (iv) deployment-style per-block execution reduces activation memory by up to $2.2\times$. 
These results point to a direction for future methods: local objectives should apply future-predictive pressure selectively across boundaries and resist shortcuts that bypass predictive content.
\end{abstract}

\section{Introduction}
\label{sec:intro}

Interest in training large language models across many loosely connected,
memory-constrained workers has grown rapidly, from low-communication
data-parallel schemes~\citep{douillard2023diloco,jaghouar2024opendiloco} to
collaborative pretraining runs spanning the
globe~\citep{jaghouar2024intellect}. The complementary axis, \emph{model
parallelism} over weak workers where no single device can hold the model,
remains harder. Pipeline-parallel training must ship both forward
activations and backward gradients across stage boundaries, and the
backward pass couples the stages' schedules and
updates~\citep{ryabinin2023swarm,ramasinghe2025subspace}.

Greedy block-wise local
learning~\citep{belilovsky2019greedy,belilovsky2020decoupled} is a
natural candidate for this setting. It splits the network into $K$
gradient-isolated blocks and trains each block through an auxiliary head
with a local loss. Forward activations still flow between stages, but no
error signal crosses a boundary. In a stage-partitioned deployment, inter-stage traffic becomes
forward-only, the backward update dependencies disappear, and each worker's
backward graph covers only one block and its auxiliary head. A line of work
has made local learning competitive with end-to-end backpropagation on
image classification~\citep{belilovsky2019greedy,pyeon2021sedona,wang2021infopro,su2024man,ma2024auglocal}.
For language, \citet{laskin2021local} trained small Transformers (6M
parameters in their main experiments) with local updates and found that
they reach a target loss with less sequential compute, but do not reach as
low a loss as backpropagation within a fixed example budget. To our
knowledge, no prior work measures how that gap behaves in autoregressive
language model (LM) pretraining at 125M--400M parameters under the standard
next-token-prediction (NTP) objective, judged by held-out likelihood, and
which auxiliary designs reduce it.

Two established accounts motivate our measurements. In restricted function
classes, joint training with upper layers corrects lower-layer feature
errors, and gradient isolation removes this \emph{backward feature
correction}~\citep{allenzhu2020backward}. On the representation side,
\citet{wang2021infopro} argue that a greedy local loss can discard
information needed downstream. We ask how these effects show up in LM
pretraining: how the likelihood gap scales with block count and model size,
and which aspects of auxiliary design reduce it.

We split 125M- and 400M-parameter Llama-style
Transformers~\citep{vaswani2017attention,touvron2023llama} into
$K\in\{1,2,4\}$ equal-depth blocks and train at Chinchilla token
budgets~\citep{hoffmann2022chinchilla}. We factor auxiliary design into the
network placed after a block, where we compare an MLP head with a
Transformer head, and the objective used to train it, where we compare NTP
with MTP. MTP targets are already present in the token stream, so they add
no inter-stage communication. Per-block losses show whether successive
blocks improve the representations they inherit, and deployment-style
measurements quantify the memory effect of isolating their backward
graphs.

Our contributions:
\begin{itemize}
  \item A token-budget-matched likelihood study of greedy local learning
  for LM pretraining across model scale and block count, taking the
  small-Transformer language-modeling results of \citet{laskin2021local}
  to 125M--400M parameters and Chinchilla-optimal budgets.
  \item Per-block evidence consistent with \emph{representation myopia}:
  later blocks barely improve their inherited representation under local
  NTP supervision.
  \item A factorized comparison showing that a Transformer-based auxiliary
  is a strong network-side intervention, while MTP is beneficial
  at the first boundary but not when added at deeper boundaries. Together
  with the NTP+CPC failure~\citep{oord2018cpc,lowe2019greedy}, these
  experiments suggest a design direction for boundary-adaptive,
  shortcut-resistant local objectives.
  \item Deployment-style memory measurements and analytical accounting of
  communication and auxiliary training compute for local learning as
  decentralized model parallelism. We release the factorized
  implementations used for these
  comparisons.\footnote{\url{https://github.com/mnmjh1215/greedy-local-learning-lm}}
\end{itemize}

\section{Related work}
\label{sec:related-main}

Greedy block-wise methods were developed primarily for image
classification. DGL established competitive locally trained
CNNs~\citep{belilovsky2019greedy,belilovsky2020decoupled}, SEDONA searches
decoupling points and auxiliary architectures~\citep{pyeon2021sedona}, and
InfoPro adds reconstruction pressure to preserve information that later
blocks need~\citep{wang2021infopro}. MAN and AugLocal copy later computation
into the local paths~\citep{su2024man,ma2024auglocal}, which in a
decentralized pipeline means extra synchronization or replicated downstream
computation. Closest to our setting, \citet{laskin2021local} study
\emph{local parallelism} on MLPs, ResNets, and small Transformer language
models. They vary the partition under one auxiliary form and report
trade-offs between compute and wall-clock time, whereas we measure the likelihood gap at
Chinchilla token budgets across scale and auxiliary design. Two recent LM
studies address different settings. LoPT locally post-trains an already
pretrained model, and its lower blocks reconstruct detached embeddings
instead of receiving an LM pretraining signal~\citep{shi2026lopt}.
DiffusionBlocks replaces likelihood training with per-block
denoising~\citep{shing2026diffusionblocks}.

Decentralized LM training spans two complementary axes. Local-update
data-parallel methods reduce synchronization between model
replicas~\citep{douillard2023diloco,jaghouar2024opendiloco,douillard2024dipaco,jaghouar2024intellect,peng2024demo}.
Model-parallel approaches instead distribute one model across unreliable or
weakly connected workers~\citep{ryabinin2023swarm,borzunov2023petals} or
compress boundary activations and gradients~\citep{ramasinghe2025subspace,janson2026subspace}.
Our study lies on the second axis, since local losses delete the backward
boundary traffic and the update dependency. Local learning can also be
combined with methods on the first. Appendix~\ref{sec:related} gives an extended review.

\section{Study design}
\label{sec:method}

\subsection{Greedy block-wise local learning}
\label{sec:method:formulation}

Consider a decoder-only Transformer with token embedding $E$, $L$
Transformer layers, and an output head (final norm and vocabulary
projection). We partition the $L$ layers into $K$ contiguous, equal-depth
blocks $F_1,\dots,F_K$. Writing $\mathrm{sg}[\cdot]$ for the stop-gradient
operator, the forward pass over a token sequence $x_{1:T}$ is
\begin{equation}
  h_0 = E(x_{1:T}), \qquad
  h_1 = F_1(h_0), \qquad
  h_k = F_k\!\left(\mathrm{sg}[h_{k-1}]\right), \quad k = 2,\dots,K,
  \label{eq:blocks}
\end{equation}
so activation values are identical to the original network, but no gradient
crosses a block boundary. Each non-final block $k < K$ is trained by an
\emph{auxiliary network} $g_k$ that maps the block output $h_k$ to a
prediction, scored by a local \emph{auxiliary loss}
$\mathcal{L}_k = \ell_k\!\left(g_k(h_k),\, x\right)$. The final block keeps
the model's real output head and the standard NTP cross-entropy
$\mathcal{L}_K$. Training minimizes $\sum_{k=1}^{K} \mathcal{L}_k$, and by
Eq.~\eqref{eq:blocks} the parameters of $(F_k, g_k)$ receive gradient from
$\mathcal{L}_k$ alone (the embedding is trained by $\mathcal{L}_1$). For
$K{=}1$ the formulation reduces exactly to standard end-to-end training.

Gradient isolation lets this formulation map onto model parallelism.
In a pipelined deployment, worker $k$ holds $(F_k,g_k)$, receives
$h_{k-1}$, and updates from $\mathcal{L}_k$ alone, so boundary traffic is
forward-only and stages have no backward update dependency
(Appendices~\ref{app:locking} and~\ref{app:comm}). Within this
formulation, each block has two design choices, the auxiliary network $g_k$ and the auxiliary objective
$\ell_k$, which the next two subsections vary.

\subsection{Auxiliary networks: \mlpaux{} and \tfaux{}}
\label{sec:method:auxnet}

We first vary $g_k$ while holding the local objective fixed to NTP.
\mlpaux{} applies a pre-normalized SwiGLU FFN with a residual connection,
followed by RMSNorm and a vocabulary projection. \tfaux{} prepends one
Transformer block to the same head, which adds local self-attention and
depth. Both heads are untied across blocks, trained only through their
local NTP losses, and discarded at inference.

\subsection{Auxiliary objective: multi-token prediction}
\label{sec:method:auxloss}

We next vary $\ell_k$ using MTP~\citep{gloeckle2024mtp,deepseekv3}. An
MTP-equipped auxiliary head first applies the same Transformer block as
\tfaux{}. Its output $u_k$ then enters an offset-specific residual FFN,
\begin{equation}
 z_{k,j}=u_k+\operatorname{FFN}_{k,j}
 \bigl(\operatorname{RMSNorm}_{k,j}(u_k)\bigr),
 \qquad j\in\{1,2\},
\end{equation}
where $j=1$ predicts $x_{t+1}$ and $j=2$ predicts $x_{t+2}$. The branches
share the final RMSNorm and vocabulary projection, following the shared-head
design of DeepSeek-V3. The $j=1$ branch is wiring-identical to \tfaux{},
so removing the $j=2$ branch recovers that control exactly. The added
branch costs one FFN and norm in parameters and, in training compute, a
second pass through the shared vocabulary projection.

We compare \emph{first-block} MTP (\tfmtp{}), which applies equally weighted
two-offset prediction only at the first auxiliary block, with
\emph{all-block} MTP, which applies a depth-weighted two-offset head at every
auxiliary block (Appendix~\ref{app:setup}). The two coincide at $K{=}2$.
In both, the final block remains NTP-only, and the shifted token targets
add no inter-stage communication.

\subsection{Experimental setting}
\label{sec:experiments:setting}

\textbf{Models and data.} We train Llama-architecture~\citep{touvron2023llama}
decoder-only
Transformers at two sizes: a 125M-class model ($d_{\mathrm{model}}{=}768$,
12 layers, 12 heads; 85M non-embedding parameters) and a 400M-class model
($d_{\mathrm{model}}{=}1024$, 24 layers, 16 heads; 308M non-embedding
parameters), with the Llama-2 32k SentencePiece
tokenizer~\citep{touvron2023llama2}. We use the
FineWeb-Edu 10BT sample~\citep{penedo2024fineweb}, packed into sequences of
4{,}096 tokens.

\textbf{Training.} All runs use a global batch of $2^{19}$ ($\approx$524k)
tokens per step and strict Chinchilla-optimal budgets of 20 tokens per
total parameter (embeddings included, counted at $K{=}1$): 5{,}100 steps
($\approx$2.7B tokens) for the 125M class and 14{,}300 steps
($\approx$7.5B tokens) for the 400M class. Hyperparameters are shared across
methods and block counts within each scale, and all quality experiments are
synchronous simulations with the same seed. Each method is trained with
$K\in\{1,2,4\}$ equal-depth blocks, and the $K{=}1$ configuration, which is
bit-identical to standard end-to-end training, serves as the baseline.
Optimizer, schedule, precision, and checkpoint details are in
Appendix~\ref{app:setup}.

\textbf{Evaluation.} The headline metric is held-out validation NLL per
token on FineWeb-Edu, always computed through the final block's real NTP
head, so it is directly comparable across methods, block counts, and to
end-to-end training.

\section{Observations}
\label{sec:experiments}

\subsection{Observation 1: the isolation gap grows with block count}
\label{sec:experiments:main}

\begin{figure}[t]
  \centering
  \includegraphics[width=\linewidth]{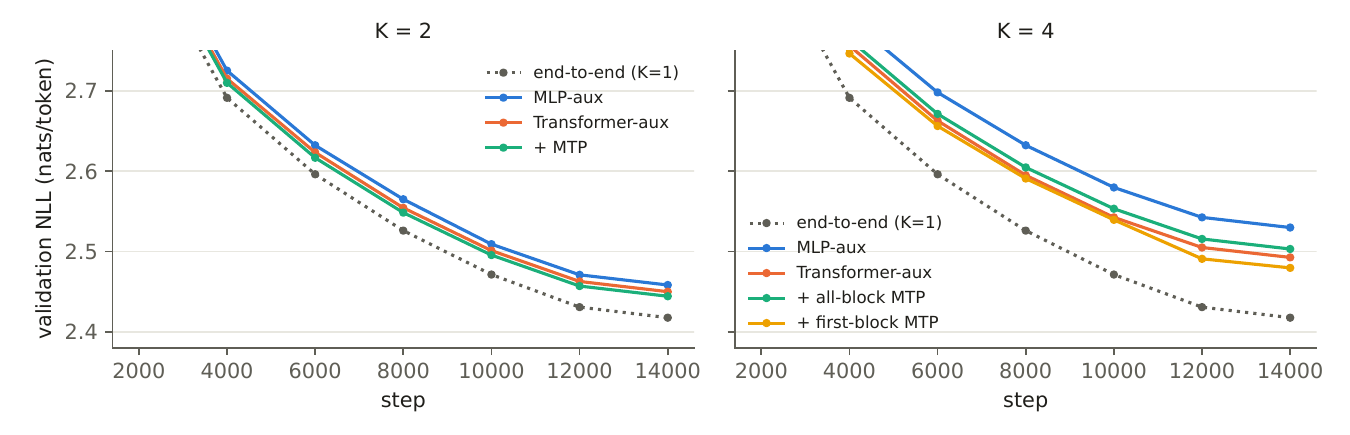}
  \caption{Validation NLL at 400M for $K{=}2$ (left) and $K{=}4$ (right).
  At $K{=}4$ the first-block placement improves over both \tfaux{} and the
  all-block MTP weighting throughout training.}
  \label{fig:val400m}
\end{figure}

Table~\ref{tab:main} compares the validation NLL of the three auxiliary
designs with the shared end-to-end baseline at both scales, and
Figure~\ref{fig:val400m} shows the corresponding 400M validation curves.

\begin{table}[t]
  \caption{Validation NLL, absolute for end-to-end ($K{=}1$) and as
  $\Delta$ against it for $K\in\{2,4\}$ (nats/token; lower is better).
  \tfmtp{} denotes first-block MTP; the all-block comparison is in
  Table~\ref{tab:placement}.}
  \label{tab:main}
  \vspace{0.5ex}
  \centering
  \small
  \begin{tabular}{lccc}
    \toprule
    & \mlpaux{} & \tfaux{} & \tfmtp{}\\
    \midrule
    \multicolumn{4}{l}{\emph{125M-class} \quad (end-to-end $K{=}1$: 2.771)}\\
    $K{=}2$ ($\Delta$) & $+0.046$ & $+0.028$ & $\mathbf{+0.017}$\\
    $K{=}4$ ($\Delta$) & $+0.132$ & $+0.078$ & $\mathbf{+0.074}$\\
    \midrule
    \multicolumn{4}{l}{\emph{400M-class} \quad (end-to-end $K{=}1$: 2.418)}\\
    $K{=}2$ ($\Delta$) & $+0.041$ & $+0.032$ & $\mathbf{+0.027}$\\
    $K{=}4$ ($\Delta$) & $+0.112$ & $+0.075$ & $\mathbf{+0.062}$\\
    \bottomrule
  \end{tabular}
\end{table}

For every auxiliary design, the gap more than doubles when the block count
increases from $K{=}2$ to $K{=}4$ (e.g.\ \mlpaux{} at 125M:
$+0.046 \rightarrow +0.132$), the same direction \citet{laskin2021local}
report for small Transformers. At $K{=}4$, moving from 125M to 400M reduces
the gap for all three designs (\mlpaux{} $+0.132\rightarrow+0.112$,
\tfaux{} $+0.078\rightarrow+0.075$, and first-block MTP
$+0.074\rightarrow+0.062$). This scale trend holds only for the deeper
partition, since at $K{=}2$ two of the three gaps increase slightly with
scale. At the lowest-gap operating point (first-block MTP, $K{=}2$, 400M),
gradient isolation costs $+0.027$ nats, about $1.1\%$ relative validation
NLL.

\subsection{Observation 2: local losses exhibit representation myopia}
\label{sec:experiments:diagnostics}

\begin{figure}[t]
  \centering
  \includegraphics[width=\linewidth]{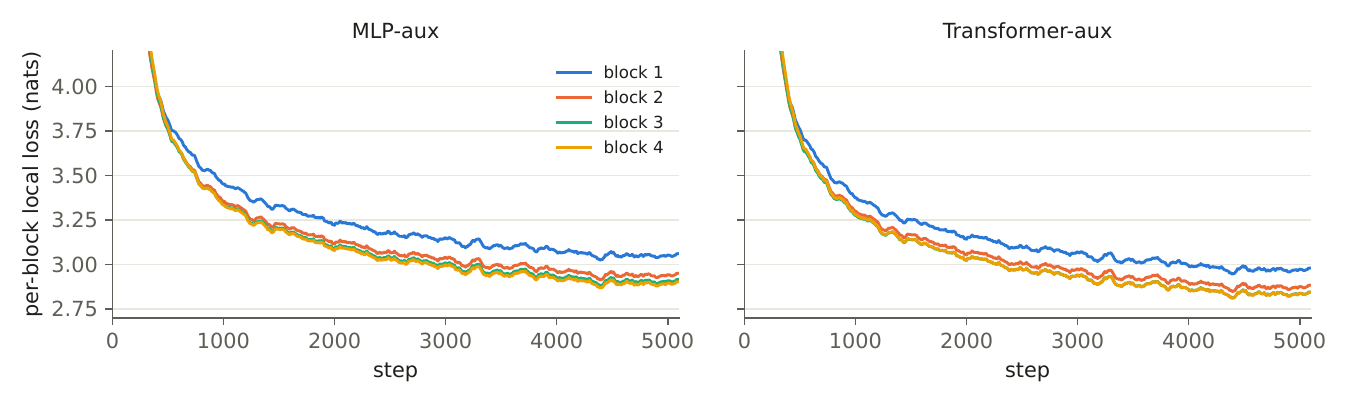}
  \caption{Per-block local losses at 125M, $K{=}4$, under \mlpaux{} (left)
  and \tfaux{} (right). The curves collapse: blocks after the first barely
  improve on the representation they inherit, and a larger head lowers all
  curves without separating them.}
  \label{fig:perblock}
\end{figure}

Figure~\ref{fig:perblock} tracks each block's local loss at 125M,
$K{=}4$. Under both NTP auxiliaries the four curves collapse onto nearly the
same trajectory. The first block already reaches almost the loss of the
full stack, and the blocks above it barely improve on the representation
they inherit, so each block behaves as a standalone next-token predictor
and does little to prepare features for its successors. We call this pattern
\emph{representation myopia}. The collapse is consistent with locally
useful features failing to improve downstream, although it does not show
directly that particular information has been discarded. The same
qualitative pattern holds at 400M (Appendix Figure~\ref{fig:perblock400}).

\subsection{Observation 3: the Transformer-based auxiliary helps most}

Replacing \mlpaux{} with \tfaux{} reduces the gap by
39\% and 41\% at 125M ($K{=}2,4$) and by 22\% and 33\% at 400M. The
per-block curves also move downward without separating, which suggests that
the stronger auxiliary makes locally predictive features easier to decode
but leaves the myopic training pattern in place. The two heads differ in
parameters and compute, so this comparison is between architectures and
does not isolate the effect of attention.

\subsection{Observation 4: MTP depends on boundary position}

Table~\ref{tab:placement} compares the two placements. At $K{=}2$, the
single auxiliary boundary is necessarily the first one, and MTP improves
over \tfaux{} by $0.011$ nats at 125M and $0.005$ at 400M. At $K{=}4$,
first-block MTP again improves over \tfaux{}, whereas equipping all
auxiliary blocks with MTP is worse than \tfaux{} at both scales. MTP
therefore helps at the first boundary, but adding the same pressure at
deeper boundaries hurts. The 400M first-block margin over \tfaux{} is
$0.013$ nats, about five times the measured ${\sim}0.0025$-nat run-to-run
noise floor, while the 125M margin ($0.004$) is only consistent in
direction. Linear and uniform all-block weighting differ by about the noise
floor (Appendix~\ref{app:mtp-ablation}).

\begin{table}[t]
  \begin{minipage}[t]{0.49\linewidth}
  \caption{MTP placement at $K{=}4$: $\Delta$ validation NLL against
  end-to-end (nats/token).}
  \label{tab:placement}
  \vspace{0.5ex}
  \centering
  \small
  \setlength{\tabcolsep}{4pt}
  \begin{tabular}{lcc}
    \toprule
    Configuration & 125M & 400M\\
    \midrule
    \tfaux{} (NTP only) & $+0.078$ & $+0.075$\\
    first-block MTP & $\mathbf{+0.074}$ & $\mathbf{+0.062}$\\
    all-block MTP, linear & $+0.082$ & $+0.086$\\
    \bottomrule
  \end{tabular}
  \end{minipage}\hfill
  \begin{minipage}[t]{0.49\linewidth}
  \caption{Per-offset training CE for all-block MTP (125M, $K{=}4$, last
  100 logged steps).}
  \label{tab:offset}
  \vspace{0.5ex}
  \centering
  \small
  \begin{tabular}{lccc}
    \toprule
    Block & $w_{t+2}$ & CE$(t+1)$ & CE$(t+2)$\\
    \midrule
    1 & $0.75$ & 3.087 & 4.588\\
    2 & $0.50$ & 2.929 & 4.500\\
    3 & $0.25$ & 2.858 & 4.518\\
    \bottomrule
  \end{tabular}
  \end{minipage}
\end{table}

The per-offset diagnostic in Table~\ref{tab:offset} shows the
corresponding trade-off inside all-block MTP. As the second-token weight
decreases across successive auxiliary blocks, next-token CE improves from
3.087 to 2.858, so the lower blocks give up more immediate predictive
sharpness under the stronger forward-looking pressure. This is consistent
with first-boundary MTP helping while the same pressure becomes
counterproductive closer to the final head. We hypothesize that early MTP
preserves broad future-predictive information that the many remaining
layers can exploit, whereas near the final head the more uncertain
second-token target competes with NTP and too little downstream depth
remains to repay the loss in immediate precision.

We also tested an NTP+CPC hybrid auxiliary at 125M, $K{=}2$. It increased
the gap to $+0.154$ nats, against $+0.028$ for \tfaux{} and $+0.017$ for
MTP, because position and document identity satisfied InfoNCE without
future-predictive content (Appendix~\ref{app:cpc}).

\subsection{Observation 5: local execution reduces memory at a training-compute cost}
\label{sec:experiments:systems}

Table~\ref{tab:memory} reports peak GPU memory per worker, measured on a
single GPU at batch $4\times4{,}096$ tokens with bf16 autocast. This is the
footprint of one pipeline stage: gradients, optimizer state, and
activations for one block plus its auxiliary head. For the MTP head we use
the sequential per-offset computation of \citet{gloeckle2024mtp}, which
runs each offset branch's forward and backward eagerly so that only one
vocabulary-sized logits tensor is alive at a time. Splitting the 400M model
into $K{=}4$ blocks cuts peak memory from 29.2 to 14.5--15.8\,GiB
(${\sim}1.9$--$2.0\times$) and activation memory from 27.7 to
12.4--13.4\,GiB (up to $2.2\times$), depending on the auxiliary head. The
Transformer-block activations scale as $1/K$, but every worker still runs a
vocabulary-sized auxiliary head, so the reduction is sublinear in $K$.
\tfmtp{} carries the MTP head only at the first stage and plain \tfaux{}
heads elsewhere, and under sequential execution its maximum-stage footprint
at $K{=}4$ is within $0.1$\,GiB of \tfaux{}. Appendix~\ref{app:setup} lists
the per-stage values.

\begin{table}[h!]
  \caption{Peak/activation GPU memory per worker (GiB; single GPU, batch
  $4\times4{,}096$, bf16). Non-E2E entries hold one block and its auxiliary
  head and report the largest stage. MTP uses sequential per-offset
  computation~\citep{gloeckle2024mtp}.}
  \label{tab:memory}
  \vspace{0.5ex}
  \centering
  \footnotesize
  \setlength{\tabcolsep}{3.5pt}
  \begin{tabular}{lcccc}
    \toprule
    & \multicolumn{2}{c}{125M-class} & \multicolumn{2}{c}{400M-class}\\
    \cmidrule(lr){2-3}\cmidrule(lr){4-5}
    Method & $K{=}2$ & $K{=}4$ & $K{=}2$ & $K{=}4$\\
    \midrule
    End-to-end ($K{=}1$) & \multicolumn{2}{c}{14.50/13.98} & \multicolumn{2}{c}{29.16/27.68}\\
    \mlpaux{}  & 11.03/10.40 & 9.52/8.56 & 19.07/17.43 & 14.52/12.41\\
    \tfaux{}   & 12.05/11.07 & 10.32/9.25 & 20.49/18.33 & 15.67/13.36\\
    \tfmtp{} (first block only) & 12.37/11.24 & 10.36/9.26 & 21.02/18.56 & 15.76/13.37\\
    \bottomrule
  \end{tabular}
\end{table}

\textbf{Compute overhead.} Auxiliary heads are discarded at inference but
add training compute, which our token-matched comparisons do not charge
(Appendix~\ref{app:compute}). One head costs $19\%$, $26\%$, and $45\%$ of
end-to-end training FLOPs at 125M for \mlpaux{}, \tfaux{}, and an MTP
head, and $9\%$, $13\%$, and $22\%$ at 400M. Most of this cost is the
vocabulary projection that every local head needs, and an MTP head runs it
twice. At $K{=}4$ the totals reach $78\%$ for \tfaux{} and $97\%$ for
first-block MTP at 125M, and $39\%$ and $48\%$ at 400M.

\section{Conclusion}
\label{sec:conclusion}

We measured greedy block-wise local learning for LM pretraining at
Chinchilla budgets. The gap to end-to-end training more than doubles from
two to four blocks, although at $K{=}4$ it shrinks with scale. Collapsed
per-block losses are consistent with representation myopia. The
Transformer-based auxiliary is an effective network-side intervention. MTP
helps at the first boundary but hurts when added at deeper ones, which makes
first-block MTP the best $K{=}4$ configuration at both scales. Finally,
per-block execution reduces activation memory by up to $2.2\times$ at an
added training-compute cost.

\textbf{Limitations.} Our scale is capped at 400M parameters, each configuration
uses a single seed, and the comparisons are token-matched rather than compute-matched. We
report measurements and leave a concrete method to future work. The
contrast between boundary-selective MTP and shortcut-prone NTP+CPC suggests
that future local objectives should preserve future-predictive content,
resist position and document-identity shortcuts, and adapt their pressure
to the position of each block boundary.

\clearpage\label{bib:start}
{\small
\bibliographystyle{plainnat}
\bibliography{ref}
}

\appendix
\raggedbottom

\section{Extended related work}
\label{sec:related}

\subsection{Greedy block-wise local learning}
\label{sec:related:local}

Greedy block-wise local learning has mainly been studied in image
classification. Decoupled greedy learning
(DGL)~\citep{belilovsky2019greedy,belilovsky2020decoupled} established that
CNNs split into gradient-isolated, locally supervised blocks can approach
end-to-end accuracy on ImageNet. SEDONA~\citep{pyeon2021sedona} searches
over decoupling points and auxiliary architectures and finds the split
decision more critical than the auxiliary choice and ``finding
appropriate lower blocks'' key to the success of greedy block-wise
learning. InfoPro~\citep{wang2021infopro} adds a reconstruction term to
each block's local task loss so that forwarded features retain input
information useful to later blocks. MAN~\citep{su2024man} adds a momentum
(EMA) copy of the next block's first layer to each block's local path, and
AugLocal~\citep{ma2024auglocal} builds each layer's auxiliary network from
a uniformly sampled, depth-reduced subset of its subsequent layers.

Two lines of work explain the performance gap between end-to-end
backpropagation and local learning. \citet{allenzhu2020backward} prove, for
restricted function classes, that layerwise training cannot match
end-to-end learning because lower-layer feature errors are corrected only
through joint training with higher layers (\emph{backward feature
correction}). InfoPro offers a complementary account on the representation
side, in which a greedy task loss can discard information needed downstream,
and supports it with decoder-based mutual-information estimates on CNN
classifiers.

\citet{laskin2021local} is the most direct precedent for local updates in
language modeling. They study \emph{local parallelism} in three forms:
greedy updates with one layer per block, \emph{chunked} updates with
several consecutive layers per block (our $K$-block partition), and
\emph{overlapping} updates, in which adjacent blocks share a boundary layer
and average its two updates. Each block is trained by an auxiliary
objective of the same form as the global one, and for their Transformer
the auxiliary classifier is a layer norm and a vocabulary projection.
They trace Pareto frontiers of compute against wall-clock time for MLPs on CIFAR-10, ResNet-50 on ImageNet, and
Transformers on LM1B, all trained for a fixed number of examples. On a
4-layer, 6M-parameter Transformer, local methods reach a loose loss target
with about $4\times$ less sequential compute and tighter targets with about
$2\times$ less, the latter only with overlapping updates. They do not reach
the same minimum within the example budget: the best validation losses are
3.62 for backpropagation, 3.66 for overlapping, 3.73 for two blocks, and
3.76 for greedy updates. A larger 6-layer, 512-wide Transformer in their
appendix gives 3.62 for backpropagation, 3.65 for two blocks, 3.68 for
three blocks, 3.68 for overlapping, and 3.80 for greedy updates.

\citet{laskin2021local} also show on ResNets that larger blocks generalize better, that locally trained
networks fit random labels worse than backpropagation does, and that
chunked local updates raise ImageNet throughput over pipelined
backpropagation by 8--41\% on real hardware despite the added auxiliary
computation.

Our study complements theirs. We train at 125M--400M parameters under
Chinchilla token budgets and vary the auxiliary network and objective,
which their study holds fixed. We therefore measure the likelihood gap as
a function of $K$, scale, and auxiliary design, whereas they measure the
frontier of compute against wall-clock time. For non-overlapping partitions, their loss gap
grows from two to three blocks in both Transformers, in agreement with our
Observation~1. They also raise, as a potential downside, that fully greedy
updates ``force the layers to learn features that are only relevant to
their local objective'', and the collapsed per-block losses of
Observation~2 are consistent with that concern.

Two recent works bring block-wise training to language models in
settings different from ours. LoPT~\citep{shi2026lopt} applies local
learning to \emph{post-training}. Its last block is trained with an SFT or
RL objective, while the lower blocks reconstruct detached input
embeddings. This presupposes a pretrained interface: from random
initialization the reconstruction target is itself random and supplies no
language-modeling signal, so the objective does not transfer to
pretraining from scratch. DiffusionBlocks~\citep{shing2026diffusionblocks}
makes blocks fully independent by reinterpreting the residual stack as
reverse diffusion and \emph{replacing} the loss with per-block L2
denoising. The converted model is no longer a likelihood-trained LM and is
evaluated only with MAUVE and judge-model generative perplexity, so its
results are not comparable to an NLL-matched end-to-end baseline.

\subsection{Layer-wise fine-tuning of LLMs}
\label{sec:related:layerwise}

A neighboring family reduces \emph{fine-tuning} cost by training only a
subset of layers at a time. LIFT~\citep{lift2023} fine-tunes a single
Transformer layer per iteration, and LISA~\citep{pan2024lisa}
importance-samples which layers to unfreeze. Both select which parameters
to \emph{update} under the single end-to-end loss at the real head, and the
forward and backward passes still traverse every layer. They therefore
reduce fine-tuning memory, not inter-stage communication or update
locking, and they adapt pretrained models instead of pretraining from
scratch.

\subsection{Decentralized LLM pretraining}
\label{sec:related:decentralized}

Decentralized pretraining has advanced along two composable axes. The
\emph{data-parallel, local-update} family reduces synchronization
frequency or gradient traffic between model replicas:
DiLoCo~\citep{douillard2023diloco} and its open replication and
scaling~\citep{jaghouar2024opendiloco}, path composition in
DiPaCo~\citep{douillard2024dipaco}, gossip-based variants that reduce
blocking communication~\citep{koneputugodage2026gossip}, the INTELLECT-1
globally distributed 10B run~\citep{jaghouar2024intellect}, and compressed
optimizer states in DeMo~\citep{peng2024demo}.

The \emph{model-parallel} family, where our work sits, splits the model
across workers. SWARM parallelism~\citep{ryabinin2023swarm} and
Petals~\citep{borzunov2023petals} route through unreliable pipeline stages,
while Subspace Networks~\citep{ramasinghe2025subspace} and learned subspace
compression~\citep{janson2026subspace} compress inter-stage activation and
gradient traffic. \citet{ramasinghe2025subspace} report that the backward
direction is the harder one to compress because errors accumulate. Local
learning complements such compression by deleting that direction entirely,
leaving only forward traffic to compress and no backward update
dependencies. The two families also
compose, since a local-update scheme can run across replicas of a locally
trained pipeline.

\section{Training and implementation details}
\label{app:setup}

All runs use AdamW with learning rate $3\times10^{-3}$, weight decay 0.033,
gradient clipping at 1.0, and bf16 mixed precision. The learning rate follows
a cosine decay to near zero, with 300 warmup steps at 125M and 1{,}000 at
400M. Final evaluations use the last saved checkpoint (step 5{,}000 for
125M and step 14{,}000 for 400M), so all reported comparisons are
step-matched.

For all-block MTP, the $k$th auxiliary block (indexed from one) uses the
linear schedule
\begin{equation}
 \mathcal{L}_k=\frac{k}{K}\,\mathrm{CE}_{k,1}
 +\left(1-\frac{k}{K}\right)\mathrm{CE}_{k,2},
 \label{eq:mtpweights}
\end{equation}
which places more second-token weight on earlier blocks. We also evaluate
uniform weighting, with $1/2$ per offset at every auxiliary block. At
$K{=}2$ there is a single auxiliary block, so first-block MTP, all-block
linear, and all-block uniform all reduce to $1/2$ per offset.

The memory benchmark executes each stage separately. At 125M and $K{=}4$,
the per-stage peaks are 10.14/10.21/10.32/9.36\,GiB for \tfaux{} and
10.34/10.25/10.36/9.41\,GiB for first-block MTP. At 400M they are
15.24/15.43/15.67/14.47\,GiB and
15.55/15.51/15.76/14.56\,GiB, respectively. MTP uses sequential per-offset
execution, so only one vocabulary-sized logits tensor is live at a time.

\section{Per-block diagnostics at 400M}
\label{app:perblock400}

\begin{figure}[t]
  \centering
  \includegraphics[width=\linewidth]{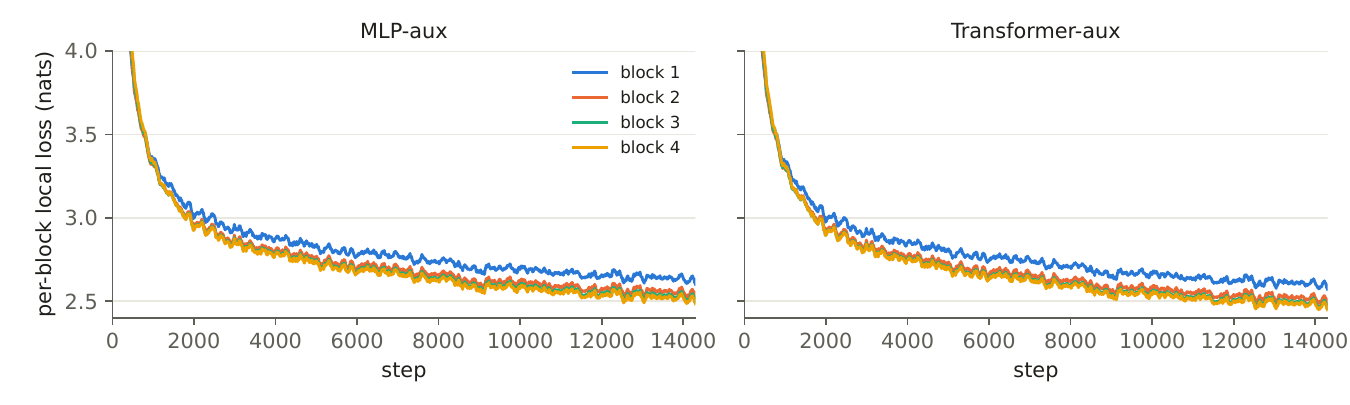}
  \caption{Per-block local losses at 400M, $K{=}4$, under \mlpaux{} (left)
  and \tfaux{} (right). As at 125M, later blocks provide only small
  improvements over the first block, and the Transformer-based auxiliary
  lowers the curves without separating them.}
  \label{fig:perblock400}
\end{figure}

Figure~\ref{fig:perblock400} repeats the $K{=}4$ per-block diagnostic at
400M. The local losses again cluster tightly after the first block under
both auxiliary networks, and the Transformer-based head shifts the curves
downward without removing their collapse, so representation myopia
persists at the larger scale.

\section{Backward and update locking}
\label{app:locking}

We follow the terminology of \citet{jaderberg2017dni}. A module is
\emph{forward locked} when it waits for its input from the preceding
module, \emph{update locked} when it waits for the loss, and
\emph{backward locked} when it waits for gradients to traverse the
downstream modules. Pipeline backpropagation therefore gates each stage's
update on a round trip through the forward and backward passes.

Local losses remove update and backward locking, because a stage can update
as soon as it has computed its own loss. Forward locking remains, since
stage $k$ still consumes $h_{k-1}$. Stages can stream successive
microbatches through this remaining forward dependency, although the usual
pipeline fill, drain, and load-imbalance effects still apply.

\section{Analytical inter-stage communication}
\label{app:comm}

Table~\ref{tab:comm} gives the analytical inter-stage traffic per token
per stage boundary for activations of width $d$ stored in $b$ bytes
($b{=}2$ for bf16). Pipeline-parallel backpropagation ships activations
forward and gradients backward ($2bd$ in total) and is update-locked by the
backward dependency. Local learning deletes the backward direction, which
halves the traffic to $bd$ per token per boundary and removes update
locking. For the 400M class ($d{=}1{,}024$, bf16), this is 2\,KiB instead
of 4\,KiB per token per boundary. Subspace
compression~\citep{ramasinghe2025subspace,janson2026subspace} projects
boundary traffic to a rank $r\ll d$ and is complementary.
\citet{ramasinghe2025subspace} report the backward direction as harder to
compress because errors accumulate, and local learning removes exactly that
direction, leaving only the easier forward one to compress. The auxiliary
objective itself adds no traffic, because MTP targets are the token stream
that every worker already holds.

\begin{table}[t]
  \caption{Analytical inter-stage communication per token per stage
  boundary ($d$: model width; $b$: bytes per element, $2$ for bf16;
  $r \ll d$: compression rank).}
  \label{tab:comm}
  \vspace{0.5ex}
  \centering
  \small
  \begin{tabular}{lcccc}
    \toprule
    Scheme & Forward & Backward & Total & Update locking\\
    \midrule
    Pipeline Parallelism (backprop) & $bd$ & $bd$ & $2bd$ & locked\\
    ~~+ Subspace Compression~\citep{ramasinghe2025subspace} & $br$ & $br$ & $2br$ & locked\\
    Local Learning & $bd$ & $0$ & $bd$ & none\\
    ~~+ Subspace Compression (fwd only) & $br$ & $0$ & $br$ & none\\
    \bottomrule
  \end{tabular}
\end{table}

\FloatBarrier
\section{Training-compute overhead of the auxiliary choices}
\label{app:compute}

Table~\ref{tab:compute} gives the analytical training FLOPs per token of
each auxiliary head, relative to end-to-end training of the same model. We
count $6$ FLOPs per token for every weight in a linear map (forward and
backward) and add the causal attention term at sequence length 4{,}096,
without counting the softmax or the embedding lookup. Auxiliary heads exist
only during training, so inference cost is unchanged, and the overhead
below is the cost a training worker pays for hosting one auxiliary
block.

Three points follow from the table. First, most of the overhead is the
vocabulary projection that every local head needs to produce a next-token
loss. The projection alone costs $16.0\%$ of end-to-end FLOPs per head at
125M and $7.1\%$ at 400M. Second, the auxiliary designs add to that term. Prepending a
Transformer block (\tfaux{} over \mlpaux{}) adds $7.0$ points per head at
125M and $3.9$ at 400M. The FFN and norm of the MTP branch add $3.1$
and $1.9$ points, but predicting a second offset requires a second pass
through the shared vocabulary projection, which adds another $16.0$ and
$7.1$ points and is the main added cost of an MTP head. First-block MTP
pays it at one head only. Compute alone does not order the results,
since all-block MTP costs the most and is worse than \tfaux{} at $K{=}4$
(Table~\ref{tab:placement}).

Third, the vocabulary size is fixed, so the projection grows only with
the model width while the end-to-end cost triples. Every overhead therefore
roughly halves from 125M to 400M. The total grows linearly with the
number of auxiliary blocks, $K{-}1$. At 125M with $K{=}4$, a \tfaux{} head
($0.24$\,GFLOP/token) costs more than the three-layer block it trains
($0.19$\,GFLOP/token), whereas at 400M the block ($0.64$) costs more than
the head ($0.35$).

All quality comparisons in this paper hold the token budget fixed, so the
gaps in Table~\ref{tab:main} are token-matched and do not charge local
learning for the overheads in Table~\ref{tab:compute}. Whether the overhead
is recovered in wall-clock time depends on the deployment, because
local updates also remove the pipeline bubbles and backward synchronization
of pipelined backpropagation. \citet{laskin2021local} report net throughput
gains for ResNets on ImageNet on real hardware despite the auxiliary
computation. We do not measure throughput.

\begin{table}[ht]
  \caption{Analytical training FLOPs of the auxiliary heads as a percentage
  of end-to-end training FLOPs per token at sequence length 4{,}096.
  End-to-end training costs $0.92$ and $2.75$\,GFLOP/token at 125M and 400M. At $K{=}2$ there
  is one auxiliary block, so the total equals the per-head value. At
  $K{=}4$ the total sums over three auxiliary blocks, and first-block MTP
  places an MTP head at the first and \tfaux{} heads at the other two.
  Totals are computed before rounding.}
  \label{tab:compute}
  \vspace{0.5ex}
  \centering
  \small
  \begin{tabular}{lcc}
    \toprule
    & 125M-class & 400M-class\\
    \midrule
    \multicolumn{3}{l}{\emph{Per auxiliary head}}\\
    ~~vocabulary projection alone & 16.0\% & 7.1\%\\
    ~~\mlpaux{} & 19.1\% & 9.0\%\\
    ~~\tfaux{} & 26.1\% & 12.9\%\\
    ~~MTP head & 45.2\% & 21.9\%\\
    \midrule
    \multicolumn{3}{l}{\emph{Total, $K{=}4$ (three auxiliary blocks)}}\\
    ~~\mlpaux{} & 57.2\% & 27.1\%\\
    ~~\tfaux{} & 78.2\% & 38.7\%\\
    ~~\tfmtp{} (first-block MTP) & 97.3\% & 47.7\%\\
    ~~all-block MTP & 135.5\% & 65.8\%\\
    \bottomrule
  \end{tabular}
\end{table}

\FloatBarrier
\section{MTP placement and weighting ablation}
\label{app:mtp-ablation}

Table~\ref{tab:mtpablation} compares the three MTP weightings defined in
Appendix~\ref{app:setup} at $K{=}4$, with \tfaux{} as the NTP-only
reference. Moving the MTP pressure from all
boundaries to the first one improves validation NLL by $0.008$ (125M) and
$0.024$ (400M) nats, whereas the schedule within all-block MTP (linear
vs.\ uniform) changes results by ${\sim}0.003$ nats, at the level of
run-to-run noise. The ablation is $K{=}4$-only by construction. At $K{=}2$
there is a single auxiliary block, so first-block and all-block placement
are the same and the linear and uniform schedules assign identical weights
($1/2$ per offset). The two $K{=}2$ runs are therefore identical
configurations that differ only in data-order noise, and their gap
($+0.017$ vs.\ $+0.020$ at 125M) gives our measured run-to-run noise floor
of ${\sim}0.0025$ nats.

\begin{table}[ht]
  \caption{MTP placement/weighting ablation at $K{=}4$: $\Delta$ validation
  NLL vs.\ end-to-end (nats/token). The all-block uniform configuration was
  not run at 400M.}
  \label{tab:mtpablation}
  \vspace{0.5ex}
  \centering
  \small
  \begin{tabular}{lcc}
    \toprule
    Configuration & 125M-class & 400M-class\\
    \midrule
    \tfaux{} (NTP only, reference) & $+0.078$ & $+0.075$\\
    first-block MTP & $\mathbf{+0.074}$ & $\mathbf{+0.062}$\\
    all-block MTP, linear ramp & $+0.082$ & $+0.086$\\
    all-block MTP, uniform & $+0.085$ & ---\\
    \bottomrule
  \end{tabular}
\end{table}

\FloatBarrier
\section{Feature-space contrastive auxiliaries are shortcut-prone}
\label{app:cpc}

Beyond token-space objectives, we tested feature-space contrastive
auxiliaries in the Greedy InfoMax/CPC
style~\citep{lowe2019greedy,oord2018cpc}: an InfoNCE loss over the block's
own future hidden states, with a Transformer-based auxiliary path and
cosine-normalized scores. Two successive shortcuts defeated it, and we
report both.

\emph{Positional shortcut.} With same-sequence negatives (pure CPC, 125M,
$K{=}2$), the InfoNCE task can be solved from position information alone,
because Transformer residual streams carry absolute position, unlike the
convolutional encoders of the CPC literature. The blocks collapsed into
content-free position encoders. The auxiliary InfoNCE reached as low as
${\sim}0.09$ nats (chance $8.3$) and was ${\sim}0.28$ at the stopping
checkpoint, while the final-head loss stalled at ${\sim}7.25$, against
${\sim}2.9$ for NTP-trained auxiliaries.

\emph{Document-identity shortcut.} We fixed the positional shortcut with
same-position, cross-sequence negatives, which lower-bound pure-position
solutions at $\log B$ for batch size $B$. A hybrid NTP+CPC auxiliary (125M,
$K{=}2$) still drove InfoNCE to ${\sim}0.5$ nats, below the
$\log B \approx 2.8$ bound, by combining position with coarse document
fingerprints. Position narrows the candidates to the target column, and the
document fingerprint picks the block's own sequence, so future-specific
content is never required. The cross-entropy anchor prevented collapse, but
the shortcut-directed pressure taxed the forwarded representation. The gap
reached $+0.154$ nats at a single cut, worse than \mlpaux{} with three cuts
($+0.132$), against $+0.028$ for \tfaux{} and $+0.017$ for the MTP
auxiliary at the same cut (Table~\ref{tab:cpc}).

\begin{table}[ht]
  \caption{Feature-space contrastive auxiliaries at 125M, $K{=}2$ (one
  cut), against the token-space auxiliaries of Table~\ref{tab:main}. The
  pure-CPC run collapsed into a position encoder and was stopped at step
  4{,}000 of 5{,}100; its auxiliary InfoNCE (${\sim}0.28$ nats at the stop)
  sits far below the same-sequence chance level while the final head never
  learns.}
  \label{tab:cpc}
  \vspace{0.5ex}
  \centering
  \small
  \begin{tabular}{lcc}
    \toprule
    Auxiliary & val NLL & $\Delta$ vs.\ end-to-end (2.771)\\
    \midrule
    CPC only & 7.33 & collapsed\\
    NTP + CPC hybrid & 2.925 & $+0.154$\\
    \midrule
    \mlpaux{} & 2.816 & $+0.046$\\
    \tfaux{} & 2.799 & $+0.028$\\
    \tfmtp{} & 2.788 & $+0.017$\\
    \bottomrule
  \end{tabular}
\end{table}

In both failures, a language model's hidden states are saturated with
identity information, namely where a token sits and which
document it came from, and InfoNCE exploits whichever of these signals the
negative set fails to control. Neither shortcut involves predicting the
future: the contrastive task is easier to win through identity
information. A negative-sampling scheme that suppresses both signals at once
might still make contrastive auxiliaries viable for language-model
pretraining, but designing one is beyond the scope of this study.

\end{document}